\documentclass[letterpaper, 10 pt, conference]{ieeeconf}  

\IEEEoverridecommandlockouts                              

\usepackage{amsmath} 
\usepackage{amssymb}  
\usepackage[]{graphicx}
\usepackage{algorithm}
\usepackage[noend]{algpseudocode}
\usepackage{cite}
\usepackage{blindtext}
\usepackage{xcolor}
\usepackage{color}
\usepackage{hyperref}

\newcommand{\vectwo}[1]{\tilde{\mathbf{#1}}}
\newcommand{\vecthree}[1]{\mathbf{#1}}
\newcommand{\R}{\mathrm{I\!R}}
\newcommand{\rot}[1]{\mathbf{R}(#1)}

\newcommand{\tplus}{{t{+}\Delta t}}
\newcommand{\vecbase}[1]{^\mathcal{B}\vectwo{#1}}
\newcommand{\W}{\mathcal{W}}
\newcommand{\B}{\mathcal{B}}

\newcommand{\cwSurface}{\textbf{cw}}
\newcommand{\ccwSurface}{\textbf{ccw}}
\newcommand{\srSurface}{\textbf{sr}}
\newcommand{\slSurface}{\textbf{sl}}
\newcommand{\cwsrEdge}{\textbf{cwsr}}
\newcommand{\cwslEdge}{\textbf{cwsl}}
\newcommand{\ccwslEdge}{\textbf{ccwsl}}
\newcommand{\ccwsrEdge}{\textbf{ccwsr}}

\title{\LARGE \bf
Polyhedral Friction Cone Estimator for Object Manipulation
}

\author{Morteza Azad, Silvia Cruciani, Michael J. Mathew, Graham Deacon and  Guillaume de Chambrier

\thanks{The authors are with the Robotics Research Team, Ocado Technology, Hatfield, UK. {\tt\small g.dechambrier@ocado.com}}
}

\begin{document}

\maketitle
\thispagestyle{empty}
\pagestyle{empty}

\begin{abstract}

A polyhedral friction cone is a set of reaction wrenches that an object can experience whilst in contact with its environment. This polyhedron is a powerful tool to control an object's motion and interaction with the environment. It can be derived analytically, upon knowledge of object and environment geometries, contact point locations and friction coefficients. We propose to estimate the polyhedral friction cone so that \textit{a priori} knowledge of these quantities is no longer required. Additionally, we introduce a solution to transform the  estimated friction cone to avoid re-estimation while the object moves. We present an analysis of the estimated polyhedral friction cone and demonstrate its application for manipulating an object in simulation and with a real robot. 

\end{abstract}

\section{INTRODUCTION}\label{introduction}
%

Non-prehensile manipulation is paramount in enlarging the set of possible interactions that a robotic manipulator can have with its environment.
When performing such manipulation tasks, exploiting external contacts enhances the system's dexterity~\cite{dafle_2014_extrinsic, eppner2015exploitation}; for instance, a wall can be used as an extension of the manipulator to achieve force closure~\cite{hou_2020_manipulation}. 

Contact-rich interactions are beneficial for applications such as the automated picking and packing of groceries~\cite{benchmarking_2020}
or other delivered goods~\cite{causo_arc}, and in-hand manipulation.
Examples of in-hand manipulation include regulating the friction applied by the gripper on the object to swing it towards a desired angle~\cite{in_hand_grav_slip_2015,sintov_2016,silvia_2017,antonova_2017_pivoting}, or slide it to a new contact point~\cite{shi_2017_dynamic_sliding,cruciani_2018_dmg}. 
Other examples exploit the interactions between object and environment to achieve a desired grasp~\cite{hou_2018_fast,nikhil_motion_cone_2020,aceituno_2020_global}, or to obtain a desired object configuration while in contact~\cite{hou_2020_manipulation,hou_2019_robust}.

Manipulation of objects while in contact requires knowledge and control of contact modes~\cite[Chapter~12]{lynch2017modern}. The reaction wrench space (RWS) representation
allows for full enumeration of all possible contact modes (sliding, sticking and separation)~\cite{huang_2020_efficient}. This representation allows the usage of the \emph{polyhedral friction cone} as a tool to analyse and control an object's motion whilst in contact~\cite{erdmann_1991}.



The polyhedral friction cone has been successfully exploited for manipulating an object and controlling its contact modes with a gripper or with the environment~\cite{hou_2020_manipulation,nikhil_motion_cone_2020,aceituno_2020_global,hou_2019_robust}.
In these works, control is within $S\!E(2)$, which implies a three-dimensional wrench space (two force and a moment components). Moreover, these works assume knowledge of:
\begin{itemize}
    \item friction coefficients;
    \item environment and object geometries;
    \item location of contact points.
\end{itemize}
This assumption is necessary for the analytical derivation of polyhedral friction cones. However, it is not always possible to obtain accurate estimates of all the above parameters, especially when robots operate in unstructured environments.

In this work, we focus on analysis and estimation of the polyhedral friction cone 
to achieve object manipulation. We relax the previously mentioned assumption by introducing an estimator that obtains the polyhedral friction cone boundary surfaces by exploring the RWS. We demonstrate, both in simulation and with real robot experiments, that the estimated friction cone can be used to control an object's contact modes and transitions between them. 

\begin{figure}
  \centering
    \includegraphics[width=0.3\columnwidth]{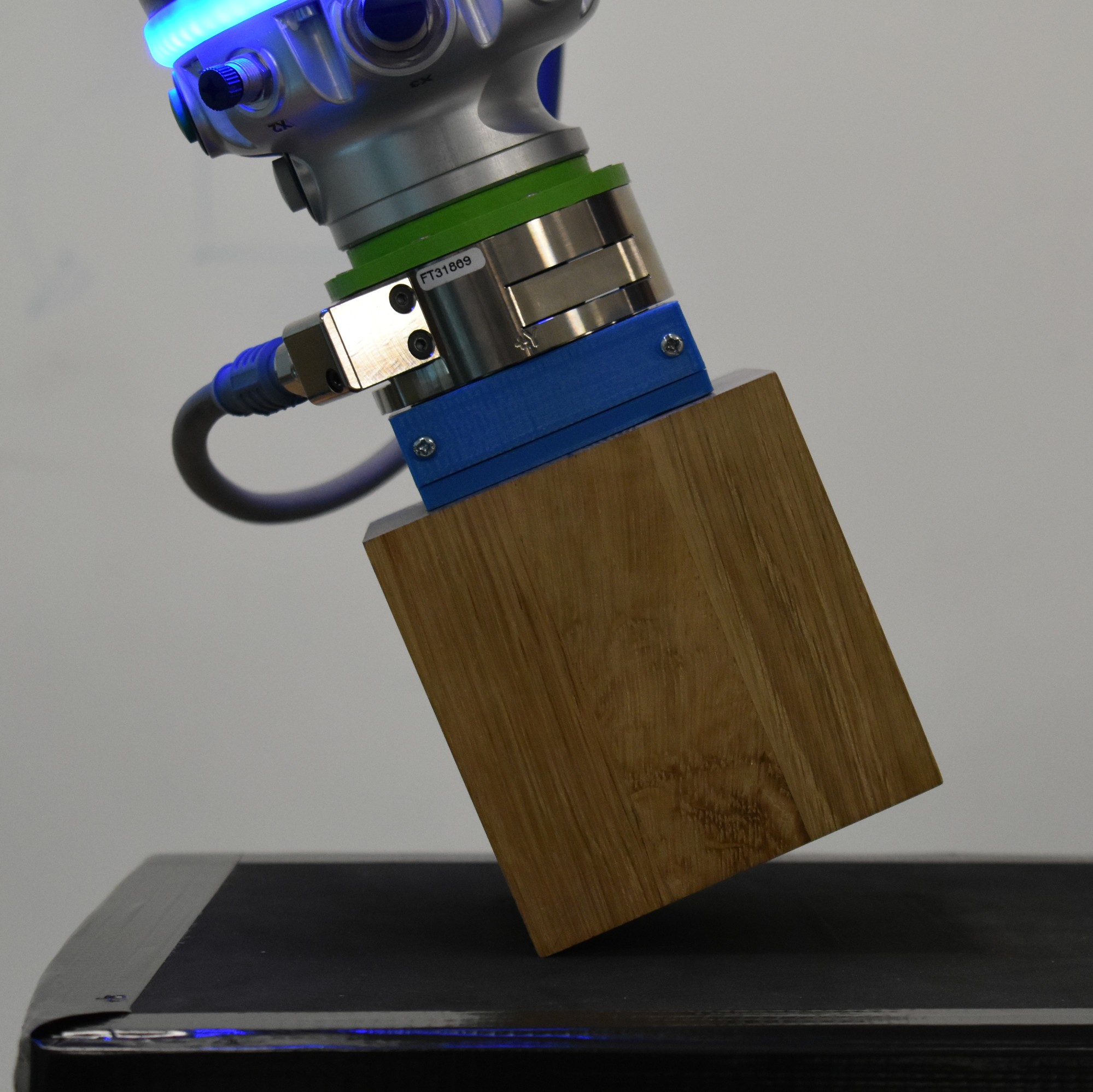}
    \includegraphics[width=0.3\columnwidth]{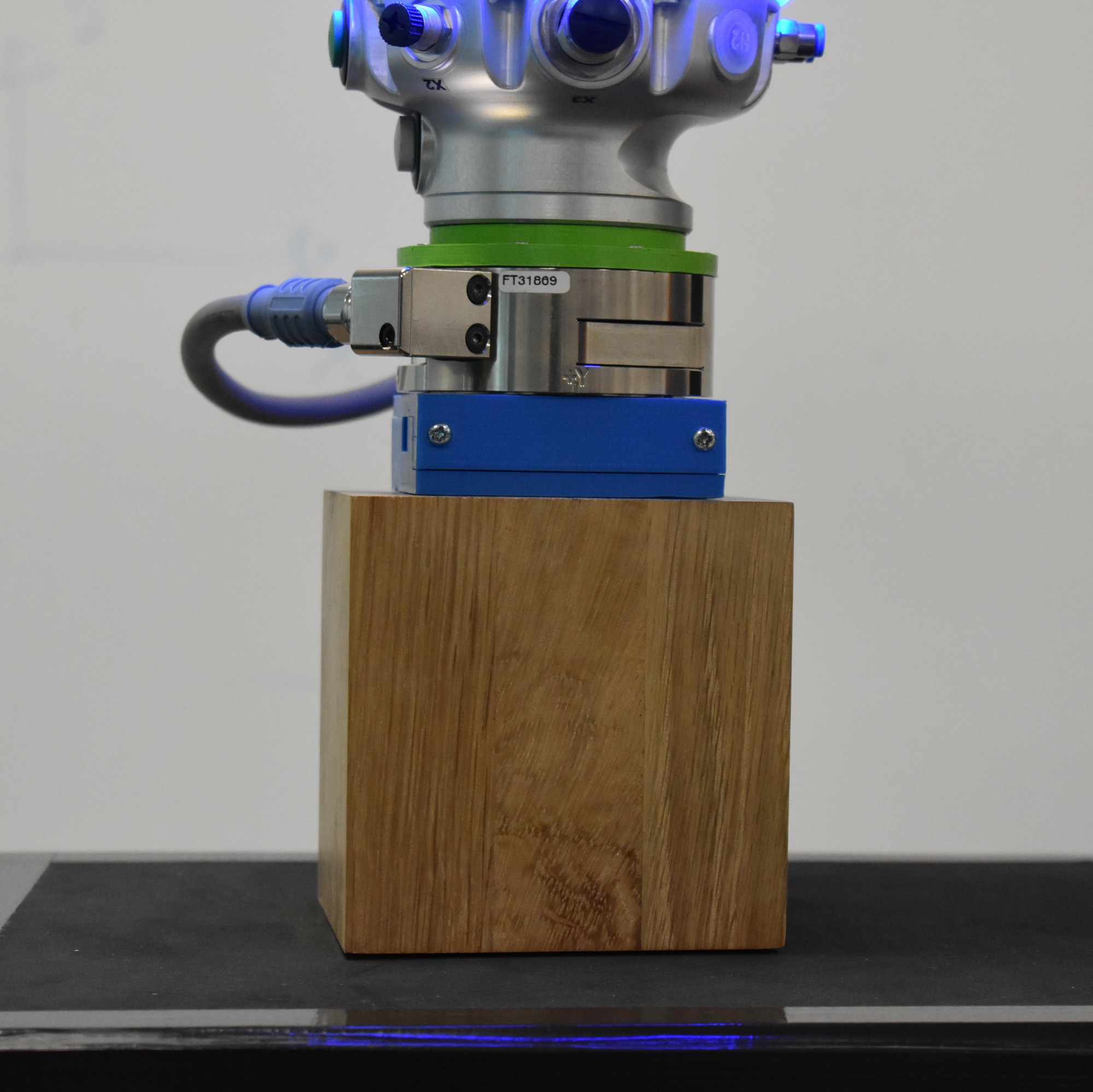}
    \includegraphics[width=0.3\columnwidth]{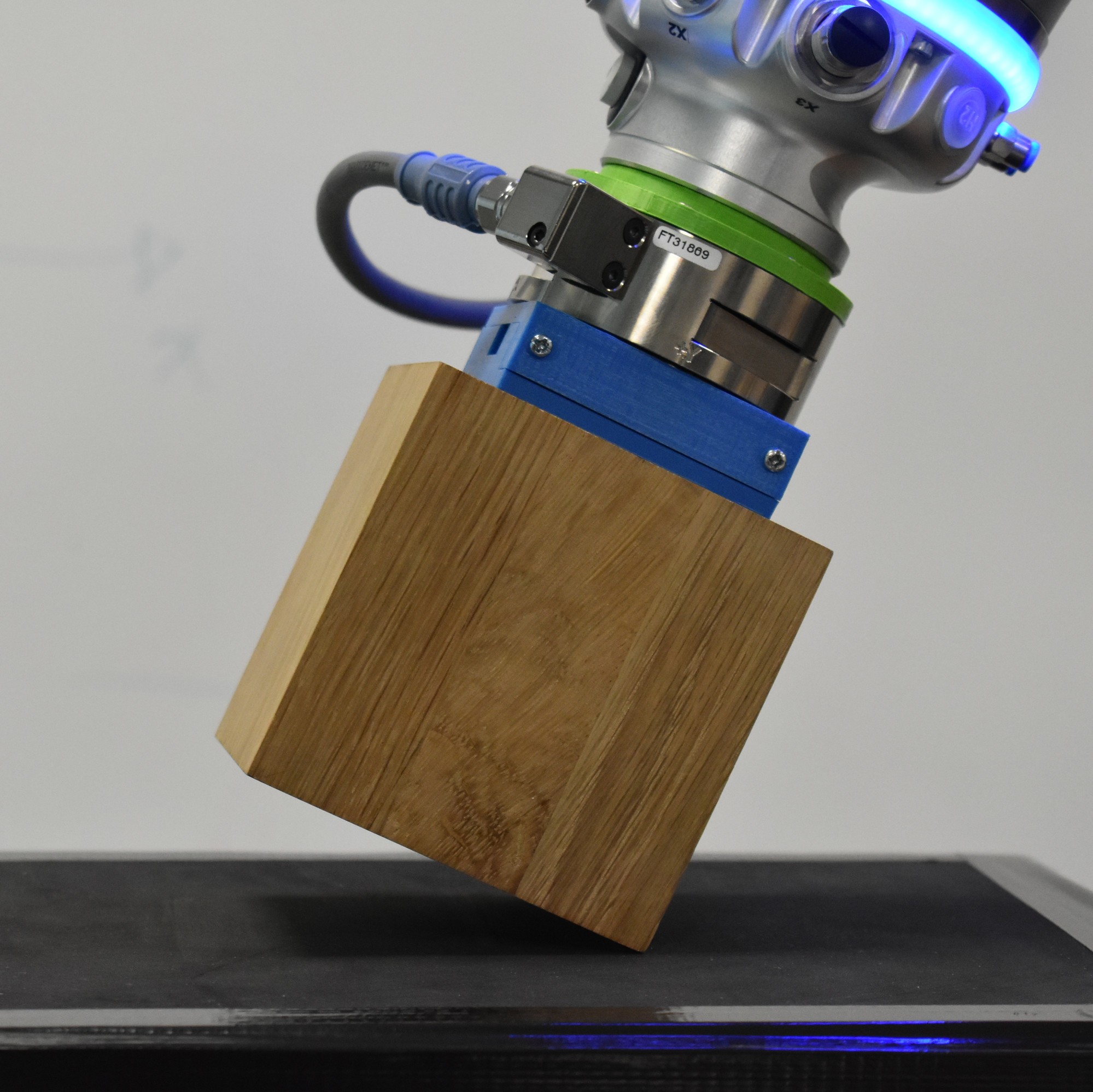}
  \caption{Example of manipulating an object in contact with the environment, by pivoting on its edge.}
  \label{fig:iiwa}
\end{figure}

We use our method to pivot an object around one of its contact points, as in Fig.~\ref{fig:iiwa}, slide it along only one contact point and slide it on the contact surface. These manipulations are executed 
without knowing the object and environment geometries, nor the location of the two initial contact points. 

\section{BACKGROUND}\label{related_work}
\subsection{Notation}
\label{subsec:notations}

In this paper, we use both 2D and 3D vectors. To distinguish between the two, we use the notation $\vectwo{v}$ for vectors in $\R^2$ and $\vecthree{v}$ for vectors in $\R^3$.
Furthermore, $\vecthree{\hat{v}}$ is a normalised vector; $\rot{\phi}$ is a $2\! \times\! 2$ rotation matrix of angle $\phi$; $\vecthree{u}\times\vecthree{v}$ is the cross product between two 3D vectors; $\vectwo{u}\times\vectwo{v}$ indicates the 3\textsuperscript{rd} component of the vector resulting from the cross product $(\vectwo{u}, 0)\times (\vectwo{v}, 0)$. 
Unless specified, all vectors are expressed in the wrench frame $\W$. We use $\vecbase{v}$ to denote a 2D vector in the base frame $\B$.

\subsection{Polyhedral Friction Cone}
When an object is in a point contact with the environment, the reaction force through this point can be represented as a wrench vector $\vecthree{w}$ in the RWS of an arbitrary reference frame~\cite[Chapter~12]{lynch2017modern}. We denote this reference frame as $\W$. Assuming a 
planar problem, the RWS lies in a three-dimensional space in which we define a wrench vector as
\begin{equation}
    \label{eq:wrench_fr_fl}
    \vecthree{w} = (\vectwo{f}, \, \vectwo{p} \times \vectwo{f}) \, ,
\end{equation}
where $\vectwo{f}$ is the reaction force and $\vectwo{p}$ is the position of the contact point, both expressed in frame $\W$.

\begin{figure}
  \centering    
  \includegraphics[width=0.75\linewidth]{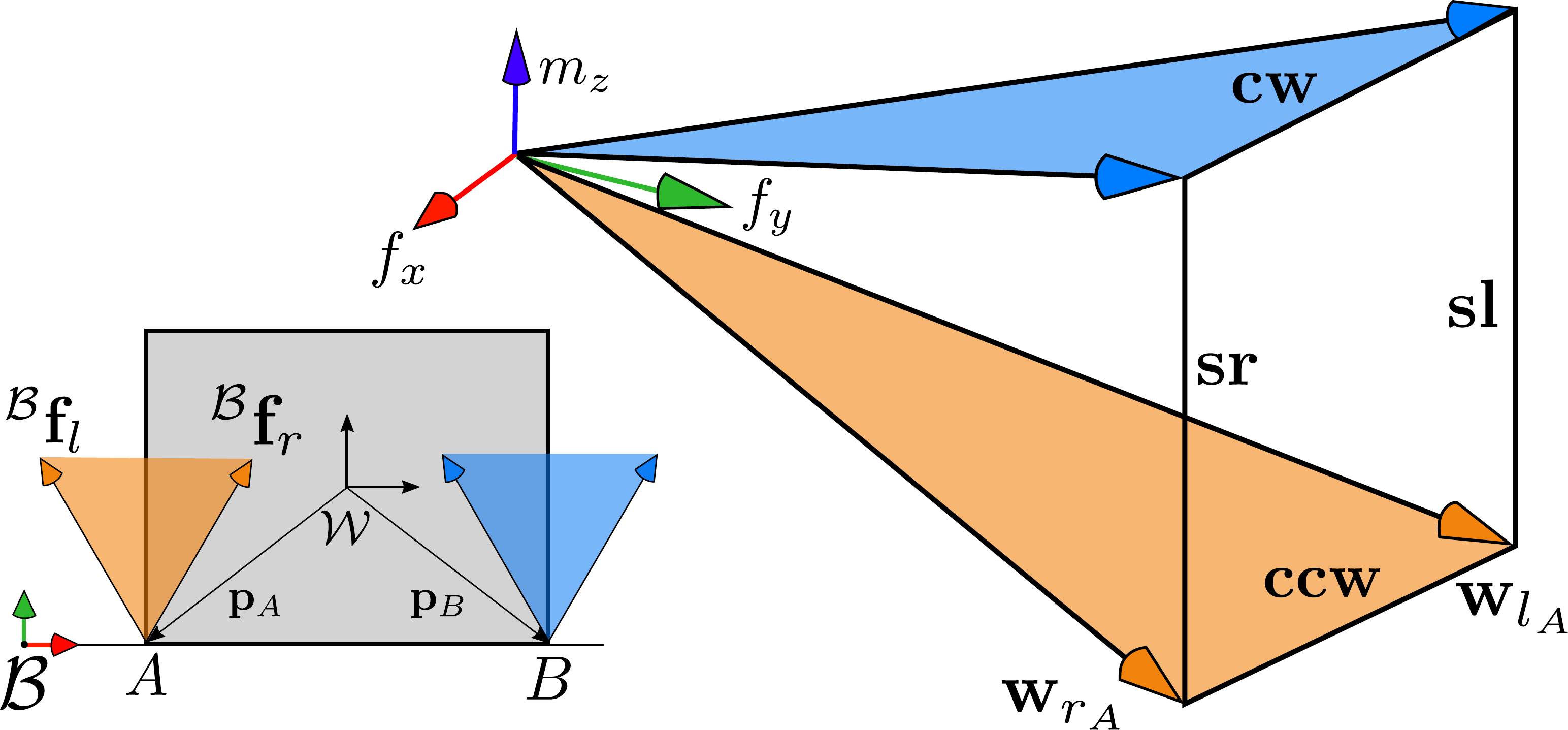}
  \caption{A rectangular object in contact at $A$ and $B$, and its corresponding polyhedral friction cone. The polyhedral fiction cone is composed of four lateral surfaces labelled \textbf{cw} (blue), \textbf{ccw} (orange), \textbf{sl} and \textbf{sr} (transparent).}
  \label{fig:block_surface_contact}
\end{figure}

By assuming Coulomb friction at the point of contact, the limits of the reaction force at this point, i.e. the limits of the friction cone, can be converted to wrench limits of the reaction wrench in the RWS by using (\ref{eq:wrench_fr_fl}). In case of multiple contact points, these limits define a polyhedral convex shape in the RWS ~\cite{erdmann_1994, mason_1991_two, graham_1997}. One of the names this convex shape, amongst many others, is the \emph{polyhedral friction cone}, which we adopt throughout this paper.
A polyhedral friction cone is shown in Fig.~\ref{fig:block_surface_contact} for a rectangular object on a line contact, approximated with contact points denoted by $A$ and $B$.

The left and right edges of both fiction cones in the base frame $\vecbase{f}_l$ and $\vecbase{f}_r$ when converted to frame $\W$ , using (\ref{eq:wrench_fr_fl}), are denoted by $\vecthree{w}_l$ and $\vecthree{w}_r$. Note that the $\vectwo{f}$ component of $\vecthree{w}$ is computed via
\begin{equation}
    \vectwo{f} = \mathbf{R}^T(\phi) \, \vecbase{f},
    \label{eq:force_transform}
\end{equation}
where $\phi$ is the angle of rotation of frame $\W$ with respect to the base frame.

\subsection{Related Work}

An early initial mention of the polyhedral friction cone appears in~\cite{erdmann_1991}, in which the author generalises the single contact friction cone to the multiple contact case. Since then, it has been widely studied and has been given many names: 
\textit{composite wrench cone} \cite[Chapter~12]{lynch2017modern}; \textit{contact wrench set} \cite[Chapter~4]{day_2016};
\textit{configuration space friction cone} \cite{erdmann_1991}; \textit{wrench space composite friction cone} \cite[Chapter~4]{graham_1997}; \emph{polyhedral convex cone} \cite{hirai1993kinematics}.
It has been used as a quality metric for grasping~\cite{gsp_2016} and for control and planning in locomotion~\cite{zmq_2004,application_feasible_metric_2017}. We review a subset of recent works in manipulation that 
either use the polyhedral friction cone directly or elements of it.

In \cite{hou_2019_robust} the authors introduce a Hybrid Force-Velocity Controller (HFVC) that can avoid unwanted
transitions between contact modes by introducing friction cone constraints. The HFVC was further used in a 
\emph{shared grasping} application~\cite{hou_2020_manipulation} where a robot manipulator rotates a non-rigidly 
attached wooden block on its sides and corners. The commanded wrench of the HFVC was obtained through 
a \emph{wrench stamping} method which selects and transforms a wrench from the polyhedral friction cone such that 
it is the most robust against changes in the contact mode (slippage or separation). 
Instead of using the polyhedron directly in a controller,
friction cones have been incorporated as constraints in planning problems such as updating a nominal trajectory to guarantee the presence of different contact modes~\cite{aceituno_2020_global}. 

Although not a polyhedral friction cone, in~\cite{nikhil_motion_cone_2020}, the authors exploit the \emph{motion cone}~\cite{mason_1986_mechanics} to plan a set of frictional pushes that move an object inside a parallel gripper to a desired configuration. A motion cone is a set of feasible motions that a rigid body can follow under the action of a frictional push. This is described as a function of the polyhedral friction cone and the gravitational wrench. The core contribution is the generalisation of the motion cone to non-horizontal surfaces.

These works all assume \textit{a priori} knowledge of the quantities outlined in section~\ref{introduction}. Our aim is to alleviate the need for this knowledge to increase the applicability of polyhedral friction cones.

\section{MANIPULATION WITH POLYHEDRON}
\label{sec:manipulation}

For the object shown in Fig.~\ref{fig:block_surface_contact}, eight modes of motion are possible: sliding left (\slSurface{}), sliding right (\srSurface{}), pivoting counter-clockwise about $A$ (\ccwSurface{}), pivoting clockwise about $B$ (\cwSurface{}), pivoting counter-clockwise while sliding left (\ccwslEdge{}) or right (\ccwsrEdge{}) and pivoting clockwise while sliding left (\cwslEdge{}) or right (\cwsrEdge{}). The first four modes are the primary modes and the remaining four are a combinations of these. The corresponding labels for the primary modes are shown in Fig.~\ref{fig:block_surface_contact}. It is known from the literature~\cite[Chapter~12]{lynch2017modern} that when undergoing primary modes of motion, the reaction wrench vector $\vecthree{w}_r$ will lie on one of the lateral surfaces of the polyhedral friction cone. For the mixed modes, $\vecthree{w}_r$ meets the lateral edges of the polyhedral friction cone, which are the blue and orange vectors shown in Figure~\ref{fig:block_surface_contact}. 

\begin{figure}
    \centering
    \includegraphics[width=0.4\textwidth]{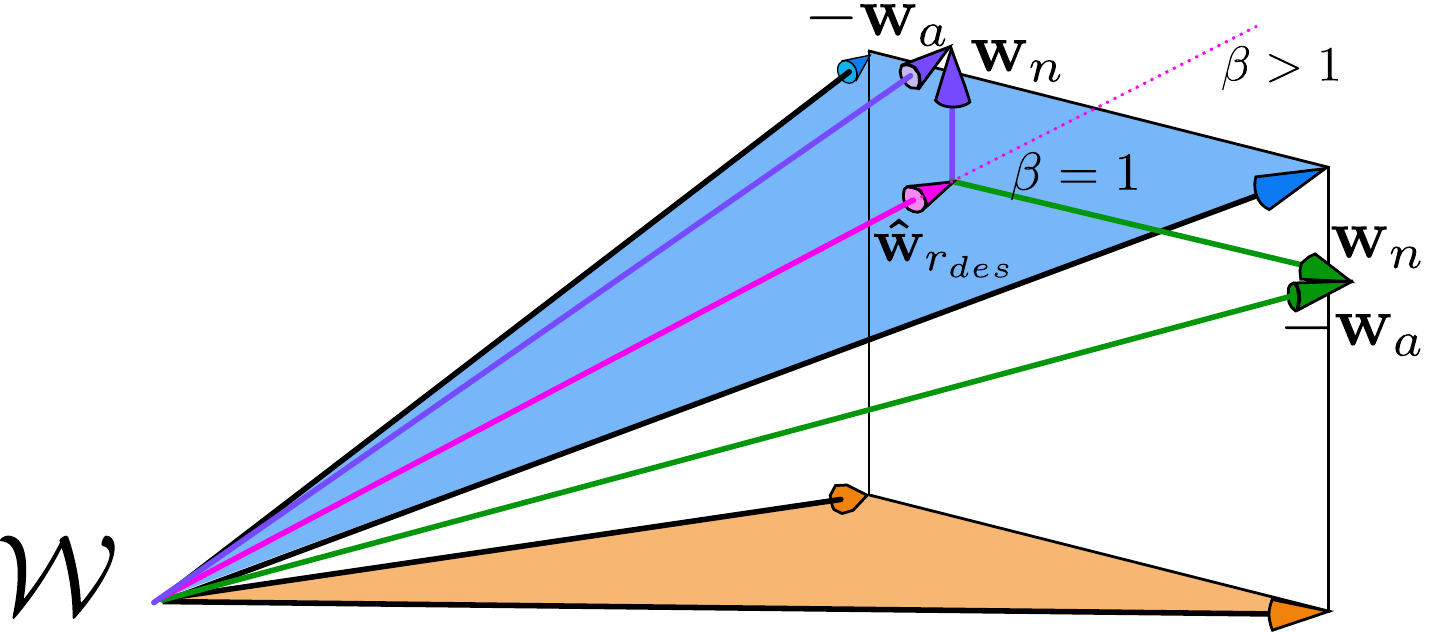}
    \caption{Control wrench for clockwise pivoting and subsequent sliding on contact point $B$. All actions start from the desired reaction wrench $\beta\vecthree{\hat{w}}_{r_{des}}$. 
    } 
    \label{fig:ctrl_polygon}
\end{figure}

If an \emph{action} wrench $\vecthree{w}_a = -\vecthree{w}_{r_{des}}$ is the negative of a defined \emph{desired wrench} $\vecthree{w}_{r_{des}}$ that lies in the middle of the polyhedron's clockwise surface, as shown in Fig.~\ref{fig:ctrl_polygon}, no motion will occur. In fact, in this case, the reaction wrench is equal and opposite to the action wrench.
If the commanded wrench changes such that $\vecthree{w}_{r_{des}}$ moves in the direction given by $\vecthree{w}_n$, see Fig.~\ref{fig:ctrl_polygon}, clockwise pivoting will start. Note that $\vecthree{w}_{r_{des}}$ cannot lie outside the polyhedron whilst contact is maintained and the difference between the action and reaction wrench will contribute to acceleration. This holds for all primary modes and, similarly, mixed modes can be achieved by having $-\vecthree{w}_a$ inside the areas which connect the surfaces of the main modes to each other.

Based on the above observation, we propose a command action wrench to follow
\begin{equation}
    \vecthree{w}_a = - (\vecthree{w}_n + \beta \hat{\vecthree{w}}_{r_{des}}) \, ,
    \label{eq:action_wrench}
\end{equation}
where $\hat{\vecthree{w}}_{r_{des}}$ is the unit vector in the direction of $\vecthree{w}_{r_{des}}$, $\beta$ is a gain for the controller and $\vecthree{w}_n$ is the wrench component in the direction of motion in the RWS. For the primary modes, this direction is 
\begin{equation}
\vecthree{w}_n = a \, \hat{\vecthree{n}} \, ,
\end{equation}
where $\hat{\vecthree{n}}$ is the unit normal to the corresponding surface, and $a$ is the magnitude of $\vecthree{w}_n$. The value of $a$ determines the magnitude of the acceleration. For the mixed modes we define
\begin{equation} \label{eq:command_wrench}
    \vecthree{w}_n = a_1 \, \hat{\vecthree{n}}_1 + a_2 \, \hat{\vecthree{n}}_2 \, ,
\end{equation}
where $\hat{\vecthree{n}}_1$ and $\hat{\vecthree{n}}_2$ are the normal vectors to the surfaces which meet at an edge and $a_1$ and $a_2$ are the magnitudes of the wrenches at those directions. For the mixed modes, $\hat{\vecthree{w}}_{r_{des}}$ is the target edge of the polyhedron. For the main modes, $\hat{\vecthree{w}}_{r_{des}}$ is the normalised mean of the two edges of the target surface. The mean vector is chosen in order to minimise the effects of uncertainties by staying away from the other two neighbouring modes. 

As mentioned earlier, $a$ determines the acceleration of motion. Therefore, by relating it to the error in the motion, we can reach a desired target position or angle by defining
\begin{equation}
    a = k_p \, e + k_d \, \dot{e} \, ,
    \label{eq:kp_kd}
\end{equation}
where $e$ is an error term and $k_p$ and $k_d$ are gains. In our scenarios, $e$ is either a linear or angular position error depending whether the goal is to slide or pivot.


When the object pivots on its corner (one contact point), the polyhedral friction cone becomes a triangular surface in the RWS. In this case, \slSurface{} and \srSurface{} merge into their corresponding edges. The normals to \slSurface{} and \srSurface{} become normals to the edges of the triangular surface, lying on the same plane. The green vector $\vecthree{w}_n$ in Fig.~\ref{fig:ctrl_polygon} shows the normal to \srSurface{} when the object has a single point of contact. Moreover, in this case, \cwSurface{} and \ccwSurface{} merge into one surface. However, the normals to these surfaces will point in opposite directions.

\section{POLYHEDRON ESTIMATION}

The polyhedron estimation is composed of three parts: \emph{estimation}, \emph{exploration} and \emph{transformation}. It is assumed that the object starts already in contact with the environment. The \emph{estimation} provides an estimate of the polyhedron's edges from a set of measured  reaction wrenches. The \emph{exploration} gathers reaction wrench samples to improve the estimation, as well as to \emph{label} the edges with the primary modes. The \emph{transformation} adapts the estimated edges as the object moves. The estimation and exploration procedure is summarised in Algorithm~\ref{alg:estimation}.

\begin{algorithm}[t]
	\small
	\caption{Polyhedron Estimation}
	\label{alg:estimation}
	\textbf{Input:} initial set from random wrenches $W_{init}$, access to the system\\
	\textbf{Output:} final estimate $W$
	\begin{algorithmic}[1]
	
	    
	    \State $W\gets W_{init}$
        \State $E\gets$basePolygonEdges($W_{init}$)
		\State $E_{explored}\gets\emptyset$
		\ForAll {$\vectwo{e}_{ij} \in E\setminus E_{explored}$}
		    \While {\textbf{not} system.hasMoved()}
		        \State $\vecthree{w}_a\gets-\vecthree{w}_r$ from (\ref{eq:heuristic_command})
		        \Comment{increase $\gamma$}
		        \State system.applyCommand($\vecthree{w}_a$)
		    \EndWhile
		    \State $W\gets$\Call{updateEstimate}{$W$, $\vecthree{w}_r$}
		    \State $E\gets$basePolygonEdges($W$)
		    \State $E_{explored}\gets E_{explored}\cup\{e_{ij}\}$
		\EndFor
		\State $\bar{P}\gets$normalise($W$)
		\State \textbf{return} $\bar{P}$
	    \State
		\Procedure{updateEstimate}{$W$, $\vecthree{w}$}
		\State $W^0\gets W\cup\{\vecthree{w}, \vecthree{0}\}$
		\State $H_{3D}\gets$ convexHull3D($W^0$)
		\If{$H_{3D}$ has not changed}
		    \State \textbf{return} $W$
		\EndIf
		\State $W\gets$ simplifyPolyhedron($H_{3D}\setminus\vecthree{0}$)
		
		\State \textbf{return} $W$
		\EndProcedure
	
	\end{algorithmic}
\end{algorithm}

\subsection{Estimation}
\label{sec:estimator}


We describe the polyhedral friction cone as the Minkowski sum of a set of $N$ normalised wrench vectors $P{=}\{\vecthree{\hat{w}}_1, \vecthree{\hat{w}}_2, ..., \vecthree{\hat{w}}_N\}$, branching from the origin $\vecthree{0}$ of the RWS. 
An estimate $\bar{P}$ of this polyhedron is obtained from a set of reaction wrenches $W{=}\{\vecthree{w}_1, \vecthree{w}_2, ...\}$ that do not produce movement on the object, i.e. they lie inside the polyhedron or on its surface; therefore, they all must lie inside the same convex volume.

To estimate $\bar{P}$, we compute the 3D convex hull $H_{3D}$ of the set of points $W^0{=}W\cup\{\vecthree{0}\}$. The convex hull removes all wrenches inside the polyhedron and leaves those which lie on the boundary surfaces. In addition, we filter out edges that lie on a plane defined by two other edges, as well as edges that are too close to others (up to a given threshold). The remaining edges replace all the ones in $W$, and their normalisation produces the vectors in $\bar{P}$.

From the polyhedron estimate we compute a base plane $\boldsymbol{\pi}$, defined by its normal and a position vector. The normal is obtained by averaging all the vectors in $\bar{P}$, and the position is randomly chosen from $W$. By intersecting $\boldsymbol{\pi}$ with rays directed along the vectors in $\bar{P}$, we obtain a set of vertices $\vecthree{w} \in V$ which forms the polyhedron's base convex polygon. We define $\vecthree{e} \in E$ to be the set of edges  connecting the  vertices $V$; an edge $\vecthree{e}_{ij}$ connects the vertices $\vecthree{w}_i, \vecthree{w}_j$.

\subsection{Exploration}
\label{sec:exploration}

\begin{figure}
  \centering
  \includegraphics[clip, scale=0.4]{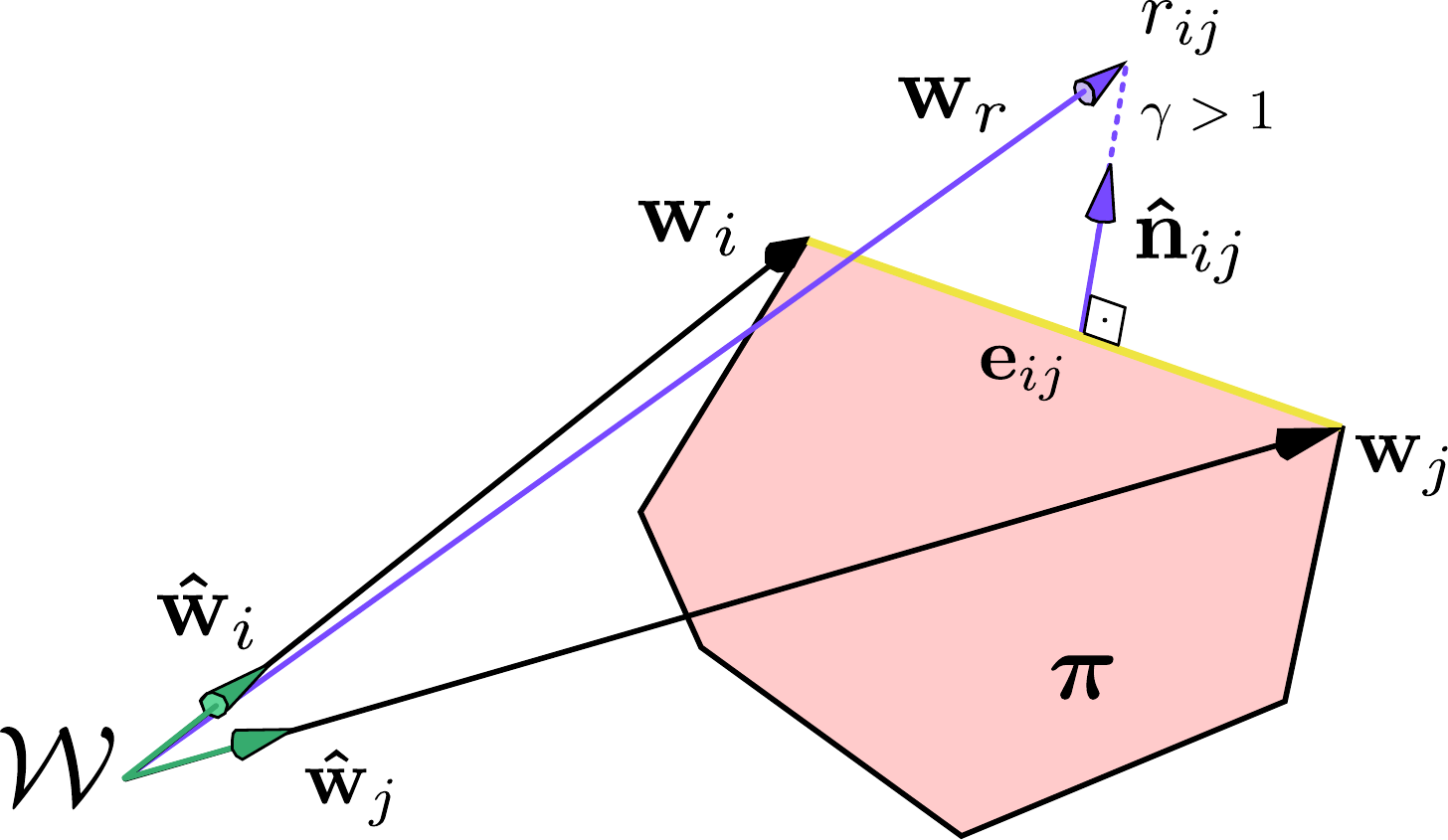}
  \caption{Estimated base polygon, with edges $\mathbf{\hat{w}}_{i}, \mathbf{\hat{w}}_{j} \in \bar{P}$ and corresponding vertices $\mathbf{w}_{i}, \mathbf{w}_{j} \in V$. The estimated base plane $\boldsymbol{\pi}$ is shown in pink. $\vecthree{e}_{ij} \in E$ is an edge selected by the search strategy, $\vecthree{\hat{n}}_{ij}$ is the normal to this edge on the plane $\boldsymbol{\pi}$, and $r_{ij}$ is the intersection of $\vecthree{w}_r$ and $\boldsymbol{\pi}$.}
  \label{fig:polygon_estimation}
\end{figure}

To begin the estimation process, $W$ is initialised by applying a number of commanded wrenches opposite to the initial measured reaction wrench, with small random perturbations. 
The goal of the search strategy is to expand $\bar{P}$ to be as close as possible to $P$ and follows these steps:
\begin{enumerate}
    \item Select the longest \emph{unexplored} edge $\vecthree{e}_{ij}\in E$. 
    \item Compute the normal $\vecthree{\hat{n}}_{ij}$ to $\vecthree{e}_{ij}$, lying on $\boldsymbol{\pi}$.
    \item Command a wrench such that the expected reaction wrench is
        \begin{equation}
        \label{eq:heuristic_command}
        \vecthree{w}_r = (\vecthree{w}_i + \vecthree{w}_j)/2 + \gamma \hat{\vecthree{n}}_{ij}, 
        \end{equation}
    where $\gamma > 0$ is a variable parameter.
\end{enumerate}
Fig.~\ref{fig:polygon_estimation} shows an example of an initial estimated convex polygon. The intersection point of $\vecthree{w}_r$ and $\boldsymbol{\pi}$ is denoted by $r_{ij}$. Starting from $\gamma{=}0$, increasing $\gamma$ will move $r_{ij}$ along $\hat{\vecthree{n}}_{ij}$, which breaks $\vecthree{e}_{ij}$ into two edges.
The gradual increment of $\gamma$ continues until motion is detected on the object, i.e. the applied wrench is on the surface of the polyhedron. Then, $\vecthree{w}_r$ is added to $W$ and $r_{ij}$ is recorded as a point on the polyhedron's surface.

The exploration continues by restarting from step 1, and selecting a new edge to explore. If motion is detected when $\gamma{=}0$, then the selected edge is already on the surface of the estimated polyhedron. In this case, the chosen $\vecthree{e}_{ij}$ is recorded as \emph{explored} to avoid re-selecting this edge (or any other edge in the same direction) in the future. The search stops when there are no more unexplored edges to select from, which means that motion has been detected on all edges of the estimated polygon.

\subsection{Labelling}
\label{sec:labeling}

To control an object's motion, the edges of $\bar{P}$ must be associated with the primary modes of motion. If we consider a single planar object in line contact with the environment, the polyhedron will have four edges. In case $\bar{P}$ has more, a finalisation step reduces them to four, by choosing those that maximise the area of the base polygon. 

For each of the four polyhedron's faces a commanded wrench is applied based on  (\ref{eq:action_wrench}) until motion is detected, with the magnitude of the out-of-plane component proportional to the area of the base polygon. If the contact configuration changes due to contact loss, it means that the two edges delimiting the relevant face are associated with the same contact point, and the surface is either \textbf{cw} or \textbf{ccw}. This contact loss can be detected as a sharp change of the measured reaction wrench. Otherwise, the edges correspond to two different contact points, and the surface is either \textbf{sl} or \textbf{sr}. 
Finally, we distinguish between clockwise or counterclockwise rotation and also between sliding left or right by observing the sign of the object's velocity.

\subsection{Transformation}
\label{sec:transformation}

Once an estimated polyhedral friction cone is obtained, it can be used to manipulate the object as explained in section \ref{sec:manipulation}. However, while the object moves, the frame $\W$ moves with respect to the base frame; this motion leads to a motion of the polyhedron's edges in the RWS. To properly manipulate the object, the polyhedron must be updated accordingly.

The objective of \emph{transformation} is to compute $\hat{\vecthree{w}}_r^{\tplus}$ and $\hat{\vecthree{w}}_l^{\tplus}$, given $\hat{\vecthree{w}}_r^t$ and $\hat{\vecthree{w}}_l^t$, where $t$ is the current instance of time, $\Delta t$ is the time step and $\hat{\vecthree{w}}_r$ and $\hat{\vecthree{w}}_r$ are the edges of the corresponding pivoting surface ($\cwSurface$ or $\ccwSurface$) of $\bar{P}$ before the start of the motion.

We denote the force component of $\hat{\vecthree{w}}^t$ by $\vectwo{f}^t$. Noticing that $\vectwo{f}^t$ is derived from (\ref{eq:force_transform}), and the direction of the force component is constant in the base frame (i.e. surface normal is constant), we write the update rule as

\begin{equation}
\vectwo{f}^{t+\Delta t} = \mathbf{R}^T(\Delta \phi) \, \vectwo{f}^{t} \, ,
\end{equation}

where $\Delta \phi$ is the change of $\phi$ during $\Delta t$.
With $\vectwo{f}^{\tplus}$ being used as the force components of $\vecthree{w}^{\tplus}$, to compute its last component we rely on the following statement: as long as a contact point $\vectwo{p}$ does not move with respect to the object, the two edges associated with it are confined on the same plane in RWS. The proof of this statement is provided in the Appendix. Therefore, the last component of $\vecthree{w}^{\tplus}$ can be derived by projecting it  onto the plane by imposing
\begin{equation}
    \label{eq:forcing_edge_on_plane}
    \hat{\vecthree{m}} \cdot \vecthree{w}^{\tplus} = 0 \, ,
\end{equation}
where $\hat{\vecthree{m}}$ is the normal to the plane computed as
\[
\hat{\vecthree{m}}{=}\frac{\hat{\vecthree{w}}_r\times\hat{\vecthree{w}}_l}{\|\hat{\vecthree{w}}_r\times\hat{\vecthree{w}}_l\|} \, .
\]
Finally, $\vecthree{w}^{\tplus}$ is normalised to obtain the transformed edge $\hat{\vecthree{w}}^{\tplus}$ of the polyhedron.

\section{EXPERIMENTS}
%

We evaluate our proposed estimation and manipulation methods by running experiments on both simulated and physical objects. The objects are cuboids that are rigidly attached to the end-effector of a KUKA IIWA14. The robot's end-effector is equipped with an ATI Delta force/torque sensor. A table in front of the robot provides the horizontal plane of contact for the object. All experiments are conducted on a plane, and an operational space controller is used to maintain a fixed reference for the remaining dimensions. 

\subsection{A Metric for Estimation Evaluation}
\label{subsec:metric}

To evaluate the accuracy of an estimated polyhedron, we define a dimensionless metric as
\begin{equation*}
    v = \frac{V_\cap}{V_\cup},
    \label{eq:metric}
\end{equation*}
where $V_\cap$ is the volume of the intersection between the estimated and ground truth polyhedrons, and $V_\cup$ is the volume of the union of the two. Note that the ground truth polyhedron can be computed analytically given the friction coefficient, the geometry of the object and the contact surface. The value of $v$ is always between $0$ and $1$. It is $0$ if the estimated polyhedron does not intersect with the ground truth and it is $1$ if the two perfectly match.

Since the polyhedral friction cones are infinite, to compute $V_\cap$ and $V_\cup$ a base plane needs to be selected. We propose to use the axis of the ground truth polyhedron as the normal to the base plane.

\subsection{Simulation Results}


Two sets of experiments were run with the drake 
simulator~\cite{drake}. The first set evaluates the results of the proposed estimation strategy detailed in section~\ref{sec:estimator}. The second set evaluates the ability of performing different contact mode transitions, as detailed in section~\ref{sec:manipulation}, with the estimated polyhedrons obtained from the first set of experiments. 


\subsubsection{Estimation}
\label{subsubsec:estimation}

\begin{figure}
  \centering
    \includegraphics[width=\columnwidth]{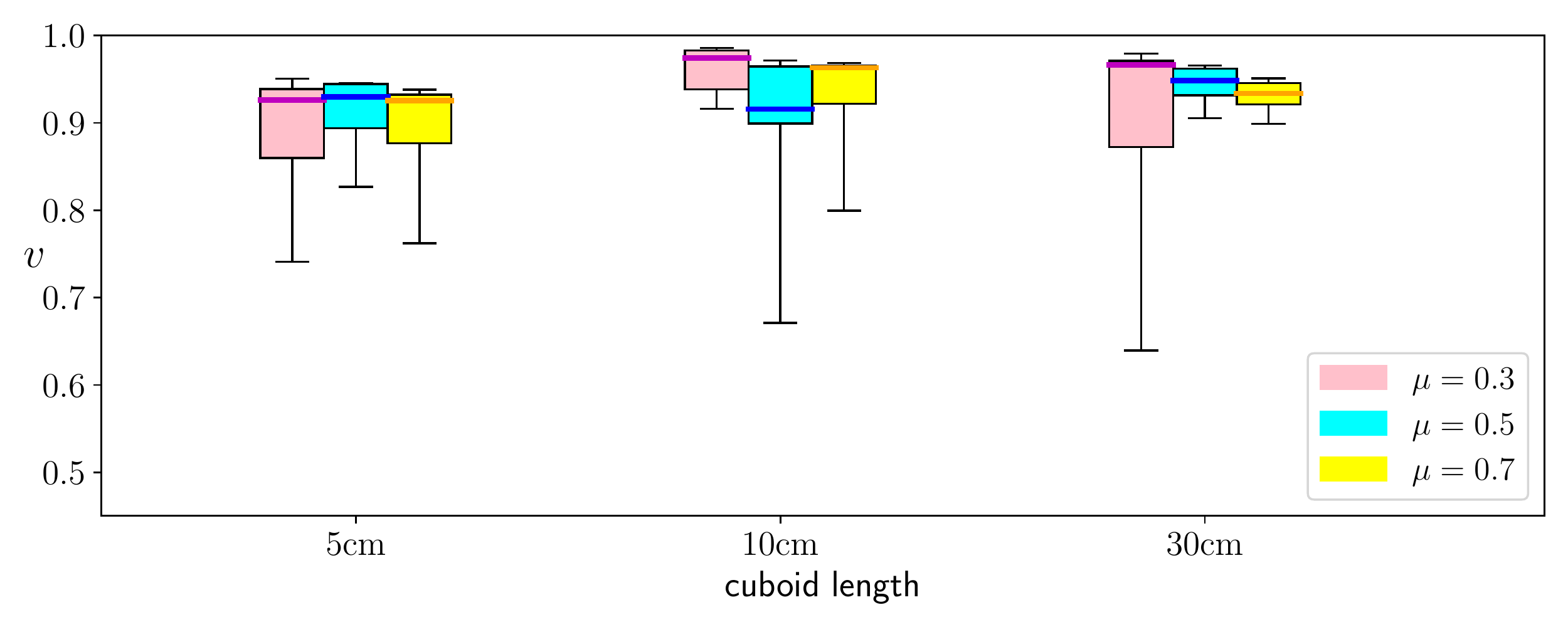}
  \caption{Box plot showing the value of $v$ with different cuboids and with different $\mu$ values on a horizontal surface. A box extends from the lower to the upper quartile values, and the thick line shows the median. The whiskers cover the whole range of data.}
  \label{fig:block_results_mu}
\end{figure}

We run the estimation algorithm in simulation for three cuboids of lengths $5$cm, $10$cm and $30$cm. 
For each cuboid, we performed experiments with three different Coulomb friction coefficients $\mu$, obtaining 10 estimates per each combination of object and $\mu$. The value of $\mu$ was set to $0.5$, $0.6$ and $0.7$. During the estimation process, we capped the maximum applied force to $10$N.
The results of these experiments are shown in Fig.~\ref{fig:block_results_mu}. The vertical axis of this plot is the metric defined in \ref{subsec:metric}.

\begin{figure}
  \centering
    \includegraphics[width=\columnwidth]{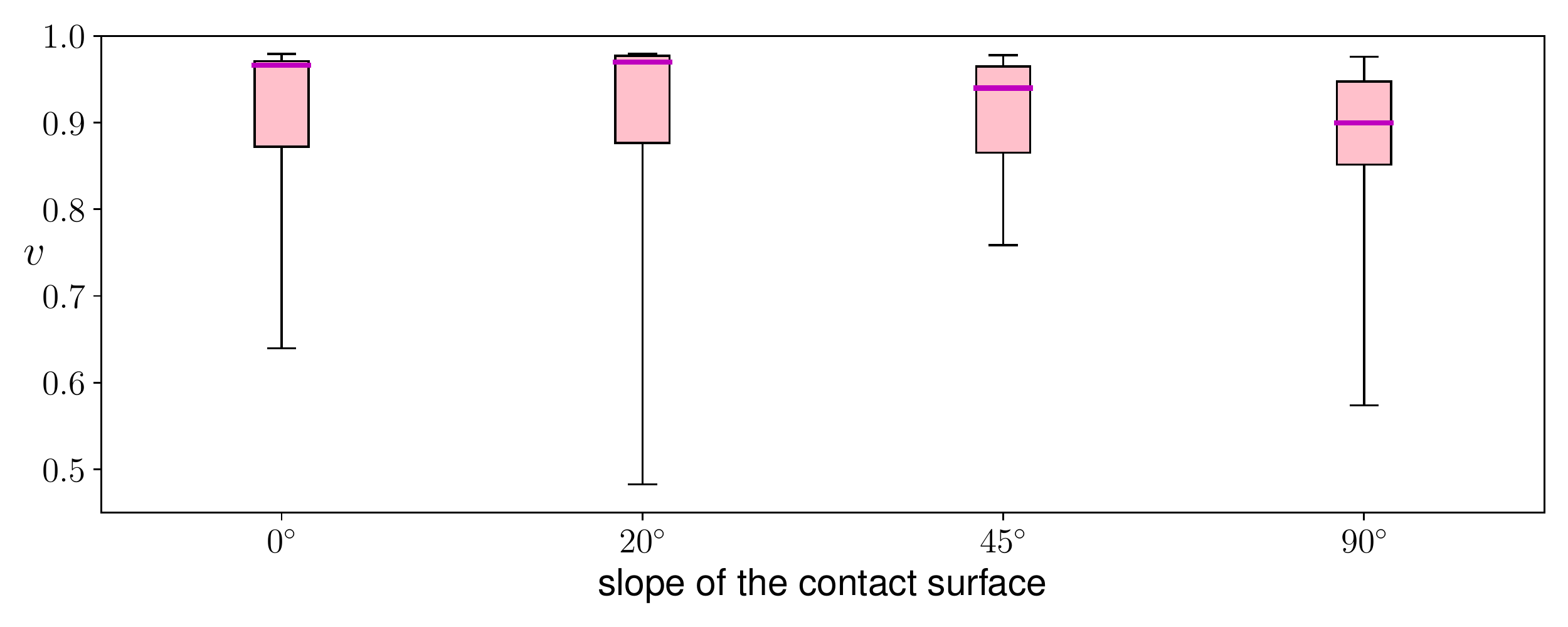}
  \caption{Box plot showing the value of $v$ for the $30$cm cuboid with different surfaces. A box extends from the lower to upper quartile values, and the thick line shows the median. The whiskers cover the whole range of data.}
  \label{fig:block_results_slopes}
\end{figure}

Fig.~\ref{fig:block_results_slopes} shows the results of estimation experiments on a surface with four different slopes, from horizontal ($0^\circ$) to vertical ($90^\circ$). The length of the used cuboid was $30$cm and $\mu$ is set to $0.3$. 10 experiments are run for each slope. Since the estimation process is independent from the geometries of the object and the environment, the results are quite robust to the change of the orientation of the contact surface.

In general, the estimated polyhedrons are often very close to the ground truth, although sometimes cover only a part of the actual polyhedron. We observed that this happens when the object suddenly slips  during exploration due to higher susceptibility to wrench variations with smaller cuboid size or low friction coefficient.

\subsubsection{Manipulation}
\label{subsubsec:manipulation}
\begin{figure*}
    \centering
    \includegraphics[width=\textwidth]{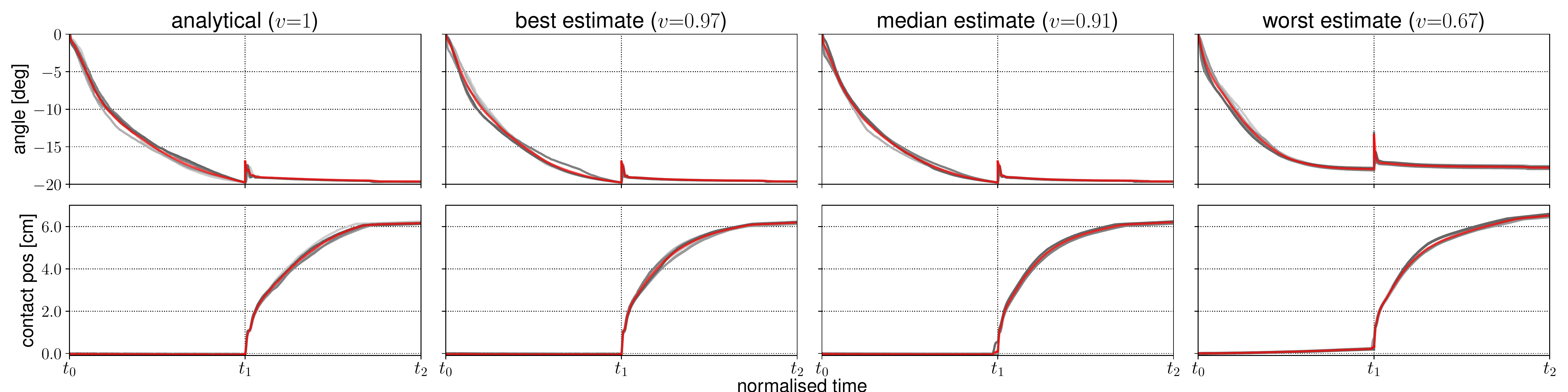}
    \caption{Plots showing the cuboid's angle and contact point position during 
    clockwise pivoting and sliding. Each column corresponds to a different polyhedron used to compute the control commands. The lines in grey show a single trial (a total of 5), and their average is shown in red. The time is normalised across the trials.}
    \label{fig:pivot_slide_results}
\end{figure*}

We evaluate the previously estimated polyhedral friction cones by observing their efficacy in accomplishing a desired manipulation task on a $10$cm long cuboid. We define the manipulation task as:
first, pivot by ${-}20^\circ$, without moving the contact point; then, slide by $6$cm, maintaining the current orientation.
The task is performed on a horizontal surface with friction coefficient $0.5$.
%
%

The polyhedron used to compute control commands are the analytical one, as well as the best, median and worst estimated polyhedral friction cones, among those obtained in the experiments detailed in \ref{subsubsec:estimation} for this configuration.

Fig.~\ref{fig:pivot_slide_results} shows the results of five trials per each polyhedron. The grey lines are for individual trials and the red ones are their averages. The values of the parameters used in (\ref{eq:kp_kd}) are $k_{p}{=}2.5$ and $k_{d}{=}0.79$; $\beta$ is set to $10$ during pivoting and $2$ during sliding. The horizontal axis shows the normalised time, when pivoting starts at $t_0$ and ends at $t_1$, and sliding starts at $t_1$ and ends at $t_2$.  The average times with the analytical polyhedron were $5.3$s for pivoting and $12.5$s for sliding. Using the best and median estimates, we observed that pivoting lasted $7.7$s and $7.6$s and sliding $19.4$s and $20.5$s, respectively. The manipulation task with the worst estimated polyhedron experienced longer average times of $19.6$s for pivoting and $30.3$s for sliding.

Both the best and median estimates allowed the pivoting task to be successful, and the object's motion is similar to the one observed with the analytical polyhedron. In contrast, the angle reached with the worst estimate is slightly above the desired one, at ${-}18^\circ$.
In the initial moments of sliding (i.e. right after $t_1$) there is a spike in the object's orientation which is due to the sudden change of mode of motion. This spike is higher when using the worst estimate of the polyhedron. However, in all cases, the controller can quickly adapt and successfully hold the desired angle.

During the pivoting phase, the contact point must remain fixed; this is successfully achieved with both the best and median estimates, which also show a successful subsequent sliding of $6$cm as desired ($6.2$cm on average). With the worst estimate, the contact point slides slightly during pivoting by $0.26$cm on average, and this reflects in a higher overall motion of $6.5$cm at the end of the execution.

These results show that the estimation process provides polyhedrons that can be used to successfully manipulate the object as desired, even if they do not perfectly match the analytical polyhedron. Even with the poorest estimates, manipulation can still be completed, albeit with small deviations from a desired behaviour.

\subsection{Real Robot Results}

\begin{figure}
    \centering
    \includegraphics[width=0.5\textwidth]{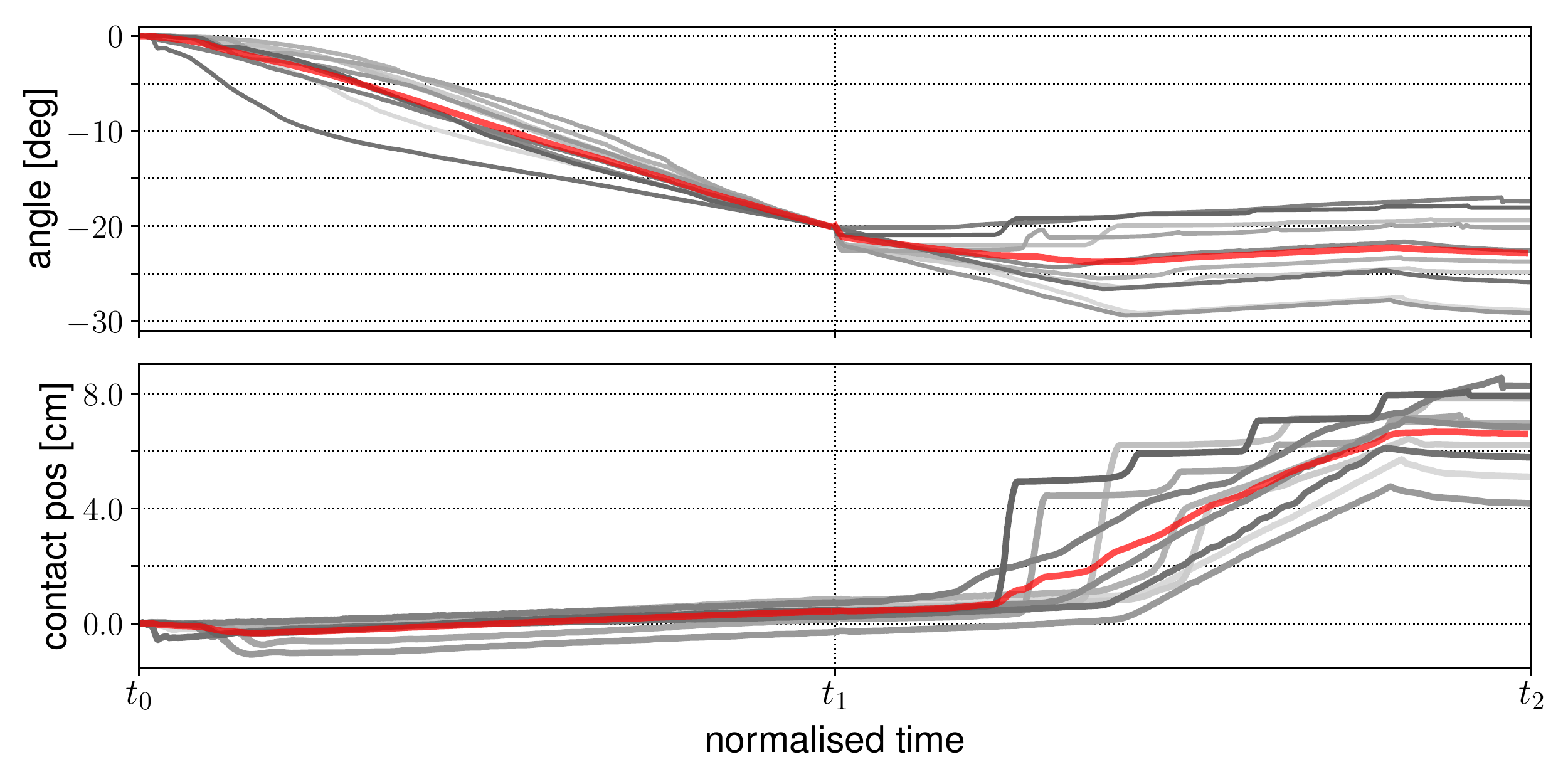}
    \caption{Plots showing the cuboid's angle and contact point position during 
    clockwise pivoting and sliding on the real robot. Each grey line is a different run, and their average is shown in red.}
    \label{fig:robot_results}
\end{figure}

For experiments on real hardware, we increased the magnitude of the maximum commanded force to $14$N. The experiments are performed using a wooden cuboid with a length of $10$cm. For safety reasons, we reset the robot's pose whenever a slippage occurred. Fig.~\ref{fig:robot_results} shows the result of pivoting and sliding motions. The target angle and sliding translation were ${-}20^\circ$ and $6$cm. We performed 10 runs, each time with a different estimated polyhedron. The gains $k_p$ and $k_d$ were set to $20$, $10$ during pivoting and $42$, $1$ during sliding. $\beta$ was set to increase from $20$ to $28$ during pivoting, to overcome stiction, and to $4$ during sliding. On average, pivoting lasted $7.2$s and sliding lasted $24.9$s. During pivoting, the object successfully reached the desired angle, but the contact point is not kept perfectly fixed and slides $3$mm on average. Sliding the wooden cube was much more challenging with the real robot compared to the simulation. Stiction causes a long delay for the object to start moving, and in many cases the observed motion is segmented in steps due to this phenomenon, and in a few instances the object does not slide as desired. Maintaining the desired angle while sliding was also challenging: in the worst instance the final angle had changed by $9^\circ$. But, the average change was less than $3^\circ$. 

The increased difficulty in applying our method on the non-simulated IIWA was due to the amount of stiction present. This caused substandard tracking between the action and reaction wrenches, and jerk. We mitigated these issues by adding extra feedback to the robot's joint torque commands, and with additional smoothing on the commanded wrench. 


\section{CONCLUSIONS}
In this paper, we presented a method that estimates and exploits the polyhedral friction cone to manipulate an object in contact with the environment, without any prior assumptions about the friction coefficients, contact point locations and geometric properties of the environment. We proposed to search in the reaction wrench space to estimate the friction polyhedron's surfaces. In addition, we proposed a labelling method for associating these surfaces with contact modes, and a transformation method for updating the polyhedron once the object starts moving.

We evaluated our solution both in simulation and with real robot experiments. We found that our mode transition method was robust with respect to estimation errors. This is a consequence of the polyhedral friction cone only partitioning the behaviour space and not modelling the motion.

In future work, instead of relying on a finalised estimate to control the object's motion, the polyhedral friction cone's surfaces can be adjusted based on the object's reaction while it is being manipulated, effectively adapting and adjusting the estimate to improve the task's execution.





\addtolength{\textheight}{-8cm}   



\section*{APPENDIX}
We note that the evolution of (\ref{eq:wrench_fr_fl}) in time entails
\begin{equation}
\label{eq:coplanarity}
\vecthree{w}^{t}\times\vecthree{w}^{\tplus}=\sin(\Delta \phi)(^\mathcal{B}\vectwo{f}\cdot ^\mathcal{B}\vectwo{f})\vecthree{p},
\end{equation}
where $\vecthree{w}^{t}$ is a wrench vector associated to the contact point $\vectwo{p}_A{=}(p_{Ax},~p_{Ay})^T$, and $\vecthree{p}{=}({-}p_{Ay},~p_{Ax},~{-}1)^T$. That is, since $\vectwo{p}_A$ is constant, an edge of the polyhedron remains always on the same plane whose normal is along $\vecthree{p}$.  Moreover, this implies that the two edges $\hat{\vecthree{w}}_r$, $\hat{\vecthree{w}}_l$ associated with the same contact point $A$ move on the same plane.

\textbf{Proof:}
with $\vecthree{u}^t{=}({{-}f^t_y,~f^t_x,~0})^T$, we can write $\vecthree{w}^t{=}\vecthree{p}\!\times\vecthree{u}^t$. Hence,
\begin{align*}
    \vecthree{w}^t\times\vecthree{w}^\tplus & = 
    (\vecthree{p}\times\vecthree{u}^t)\times(\vecthree{p}\times\vecthree{u}^\tplus) \\
    & =  ((\vecthree{p}\times\vecthree{u}^t)\cdot \vecthree{u}^\tplus)\vecthree{p}
     - ((\vecthree{p}\times\vecthree{u}^t)\cdot \vecthree{p})\vecthree{u}^\tplus \\
    & =  (\vecthree{p}\cdot(\vecthree{u}^t\times\vecthree{u}^\tplus))\vecthree{p} \\
    & =  (f^t_y f^\tplus_x - f^t_x f^\tplus_y)\vecthree{p}.
\end{align*}
Since $f^t_y f^\tplus_x {-} f^t_x f^\tplus_y{=}\vectwo{f}^\tplus\times\vectwo{f}^t$, which is equivalent to $(^\mathcal{B}\vectwo{f}\cdot^\mathcal{B}\vectwo{f})\sin(\Delta \phi)$, (\ref{eq:coplanarity}) is proved.

\section*{ACKNOWLEDGMENT}

The authors would like to thank the lab technician Tomasz Roszak for his efforts in constructing the experiment setup.


\bibliographystyle{IEEEtran}
\bibliography{BIC}

\begin{thebibliography}{10}
\providecommand{\url}[1]{#1}
\csname url@rmstyle\endcsname
\providecommand{\newblock}{\relax}
\providecommand{\bibinfo}[2]{#2}
\providecommand\BIBentrySTDinterwordspacing{\spaceskip=0pt\relax}
\providecommand\BIBentryALTinterwordstretchfactor{4}
\providecommand\BIBentryALTinterwordspacing{\spaceskip=\fontdimen2\font plus
\BIBentryALTinterwordstretchfactor\fontdimen3\font minus
  \fontdimen4\font\relax}
\providecommand\BIBforeignlanguage[2]{{%
\expandafter\ifx\csname l@#1\endcsname\relax
\typeout{** WARNING: IEEEtran.bst: No hyphenation pattern has been}%
\typeout{** loaded for the language `#1'. Using the pattern for}%
\typeout{** the default language instead.}%
\else
\language=\csname l@#1\endcsname
\fi
#2}}

\bibitem{dafle_2014_extrinsic}
N.~C. {Dafle}, A.~{Rodriguez}, R.~{Paolini}, B.~{Tang}, S.~S. {Srinivasa},
  M.~{Erdmann}, M.~T. {Mason}, I.~{Lundberg}, H.~{Staab}, and T.~{Fuhlbrigge},
  ``Extrinsic dexterity: In-hand manipulation with external forces,'' in
  \emph{2014 IEEE International Conference on Robotics and Automation (ICRA)},
  2014, pp. 1578--1585.

\bibitem{eppner2015exploitation}
C.~Eppner, R.~Deimel, J.~Alvarez-Ruiz, M.~Maertens, and O.~Brock,
  ``Exploitation of environmental constraints in human and robotic grasping,''
  \emph{The International Journal of Robotics Research}, vol.~34, no.~7, pp.
  1021--1038, 2015.

\bibitem{hou_2020_manipulation}
Y.~{Hou}, Z.~{Jia}, and M.~T. {Mason}, ``Manipulation with shared grasping,''
  in \emph{Robotics: Science and Systems (RSS)}, 2020.

\bibitem{benchmarking_2020}
H.~{Mnyusiwalla}, P.~{Triantafyllou}, P.~{Sotiropoulos}, M.~A. {Roa},
  W.~{Friedl}, A.~M. {Sundaram}, D.~{Russell}, and G.~{Deacon}, ``A bin-picking
  benchmark for systematic evaluation of robotic pick-and-place systems,''
  \emph{IEEE Robotics and Automation Letters}, vol.~5, no.~2, pp. 1389--1396,
  2020.

\bibitem{causo_arc}
A.~Causo, J.~Durham, K.~Hauser, K.~Okada, and A.~Rodriguez, \emph{Advances on
  Robotic Item Picking}.\hskip 1em plus 0.5em minus 0.4em\relax Springer
  International Publishing, 2020.

\bibitem{in_hand_grav_slip_2015}
F.~E. {Viña B.}, Y.~{Karayiannidis}, K.~{Pauwels}, C.~{Smith}, and
  D.~{Kragic}, ``In-hand manipulation using gravity and controlled slip,'' in
  \emph{2015 IEEE/RSJ International Conference on Intelligent Robots and
  Systems (IROS)}, 2015, pp. 5636--5641.

\bibitem{sintov_2016}
A.~{Sintov}, O.~{Tslil}, and A.~{Shapiro}, ``Robotic swing-up regrasping
  manipulation based on the impulse–momentum approach and clqr control,''
  \emph{IEEE Transactions on Robotics}, vol.~32, no.~5, pp. 1079--1090, 2016.

\bibitem{silvia_2017}
S.~{Cruciani} and C.~{Smith}, ``In-hand manipulation using three-stages open
  loop pivoting,'' in \emph{2017 IEEE/RSJ International Conference on
  Intelligent Robots and Systems (IROS)}, 2017, pp. 1244--1251.

\bibitem{antonova_2017_pivoting}
R.~Antonova, S.~Cruciani, C.~Smith, and D.~Kragic, ``Reinforcement learning for
  pivoting task,'' in \emph{arXiv:1703.00472}, 2017.

\bibitem{shi_2017_dynamic_sliding}
J.~{Shi}, J.~Z. {Woodruff}, P.~B. {Umbanhowar}, and K.~M. {Lynch}, ``Dynamic
  in-hand sliding manipulation,'' \emph{IEEE Transactions on Robotics},
  vol.~33, no.~4, pp. 778--795, 2017.

\bibitem{cruciani_2018_dmg}
S.~{Cruciani}, C.~{Smith}, D.~{Kragic}, and K.~{Hang}, ``Dexterous manipulation
  graphs,'' in \emph{2018 IEEE/RSJ International Conference on Intelligent
  Robots and Systems (IROS)}, 2018, pp. 2040--2047.

\bibitem{hou_2018_fast}
Y.~Hou, Z.~Jia, and M.~T. Mason, ``Fast planning for 3d any-pose-reorienting
  using pivoting,'' in \emph{2018 IEEE International Conference on Robotics and
  Automation (ICRA)}, 2018, pp. 1631--1638.

\bibitem{nikhil_motion_cone_2020}
N.~Chavan-Dafle, R.~Holladay, and A.~Rodriguez, ``Planar in-hand manipulation
  via motion cones,'' \emph{The International Journal of Robotics Research},
  vol.~39, no. 2-3, pp. 163--182, 2020.

\bibitem{aceituno_2020_global}
B.~{Aceituno-Cabezas} and A.~{Rodriguez}, ``A global quasi-dynamic model for
  contact-trajectory optimization,'' in \emph{Robotics: Science and Systems
  (RSS)}, 2020.

\bibitem{hou_2019_robust}
Y.~{Hou} and M.~T. {Mason}, ``Robust execution of contact-rich motion plans by
  hybrid force-velocity control,'' in \emph{2019 International Conference on
  Robotics and Automation (ICRA)}, 2019, pp. 1933--1939.

\bibitem{lynch2017modern}
K.~Lynch and F.~Park, \emph{Modern Robotics}.\hskip 1em plus 0.5em minus
  0.4em\relax Cambridge University Press, 2017.

\bibitem{huang_2020_efficient}
E.~Huang, X.~Cheng, and M.~T. Mason, ``Efficient contact mode enumeration in
  3d,'' in \emph{Workshop on the Algorithmic Foundations of Robotics}, 2020.

\bibitem{erdmann_1991}
M.~{Erdmann}, ``A configuration space friction cone,'' in \emph{Proceedings
  IROS '91:IEEE/RSJ International Workshop on Intelligent Robots and Systems
  '91}, 1991, pp. 455--460 vol.2.

\bibitem{erdmann_1994}
M.~Erdmann, ``On a representation of friction in configuration space,''
  \emph{The International Journal of Robotics Research}, vol.~13, no.~3, pp.
  240--271, 1994.

\bibitem{mason_1991_two}
M.~T. {Mason}, ``Two graphical methods for planar contact problems,'' in
  \emph{Proceedings IROS 'IEEE/RSJ International Workshop on Intelligent Robots
  and Systems '91}, 1991, pp. 443--448 vol.2.

\bibitem{graham_1997}
G.~E. Deacon, ``Accomplishing task-invariant assembly strategies by means of an
  inherently accommodating robot arm,'' Ph.D. dissertation, Department of
  Artificial Intelligence, University of Edinburgh, 1997.

\bibitem{day_2016}
H.~Dai, ``Robust multi-contact dynamical motion planning using contact wrench
  set,'' Ph.D. dissertation, Massachusetts Institute of Technology, Cambridge,
  {USA}, 2016.

\bibitem{hirai1993kinematics}
S.~Hirai and H.~Asada, ``Kinematics and statics of manipulation using the
  theory of polyhedral convex cones,'' \emph{The International Journal of
  Robotics Research}, vol.~12, no.~5, pp. 434--447, 1993.

\bibitem{gsp_2016}
R.~{Krug}, A.~J. {Lilienthal}, D.~{Kragic}, and Y.~{Bekiroglu}, ``Analytic
  grasp success prediction with tactile feedback,'' in \emph{2016 IEEE
  International Conference on Robotics and Automation (ICRA)}, 2016, pp.
  165--171.

\bibitem{zmq_2004}
M.~Vukobratovic and B.~Borovac, ``Zero-moment point - thirty five years of its
  life.'' \emph{I. J. Humanoid Robotics}, vol.~1, pp. 157--173, 03 2004.

\bibitem{application_feasible_metric_2017}
R.~Orsolino, M.~Focchi, C.~Mastalli, H.~Dai, D.~Caldwell, and C.~Semini,
  ``Application of wrench based feasibility analysis to the online trajectory
  optimization of legged robots,'' \emph{IEEE Robotics \& Automation Letters
  (RA-L) 2018}, vol.~PP, 12 2017.

\bibitem{mason_1986_mechanics}
M.~T. Mason, ``Mechanics and planning of manipulator pushing operations,''
  \emph{The International Journal of Robotics Research}, vol.~5, no.~3, pp.
  53--71, 1986.

\bibitem{drake}
\BIBentryALTinterwordspacing
R.~Tedrake and the Drake Development~Team, ``Drake: Model-based design and
  verification for robotics,'' 2019. [Online]. Available:
  \url{https://drake.mit.edu}
\BIBentrySTDinterwordspacing

\end{thebibliography}

\end{document}